\pdfoutput=1

\documentclass[11pt]{article}

\usepackage{acl}

\usepackage{times}
\usepackage{latexsym}

\usepackage[T1]{fontenc}

\usepackage[utf8]{inputenc}

\usepackage{microtype}

\usepackage{algorithm}
\usepackage{algorithmic}
\usepackage{CJKutf8}
\usepackage{graphicx}
\usepackage{booktabs}  
\usepackage{threeparttable}
\usepackage{multirow}
\usepackage{float} 
\usepackage{subfigure} 
\usepackage{bm}
\usepackage{amsmath}
\usepackage{makecell}
\usepackage{amsfonts}
\usepackage{hyperref}

\title{C$^3$KG: A Chinese Commonsense Conversation Knowledge Graph}

\author{Dawei Li, Yanran Li\thanks{Corresponding author: yanranli.summer@gmail.com} , Jiayi Zhang, Ke Li, Chen Wei, Jianwei Cui, Bin Wang \\
        Xiaomi AI Lab \\
        \texttt{lidawei,liyanran,zhangjiayi3,like16,} \\ 
        \texttt{weichen,cuijianwei,wangbin11@xiaomi.com}}

\begin{document}
\maketitle
\begin{abstract}
Existing commonsense knowledge bases often organize tuples in an isolated manner, which is deficient for commonsense conversational models to plan the next steps. To fill the gap, we curate a large-scale multi-turn human-written conversation corpus, and create the first Chinese commonsense conversation knowledge graph which incorporates both social commonsense knowledge and dialog flow information. To show the potential of our graph, we develop a graph-conversation matching approach, and benchmark two graph-grounded conversational tasks. Our code and data could be found in \url{https://github.com/XiaoMi/C3KG}.
\end{abstract}

\section{Introduction}
Commonsense knowledge describes facts and related judgments in our everyday world, which is essential for machine when interacting with humans. These years have witnessed a growing number of literature incorporating commonsense knowledge into various downstream tasks~\cite{bauer-etal-2018-commonsense,Chen2019IncorporatingSC,Lin2019KagNetKG,Guan2019StoryEG,ji-etal-2020-language}.

Recently, \citet{atomic} curate ATOMIC, a large-scale commonsense knowledge base, which covers event-centered social aspects of inferential knowledge tuples. For example, there exist tuples like \{\emph{PersonX adopts a cat}, \verb|xEffect|, \emph{feels happy}\} and \{\emph{PersonX adopts a cat}, \verb|xWant|, \emph{company}\}. Here, \verb|xEffect| and \verb|xWant| are two of nine relations defined in ATOMIC to infer people's mental states for a given event, e.g., \emph{PersonX adopts a cat}. As such, it is promising to detect ATOMIC events mentioned in conversations, and utilize the inferred knowledge when developing social chatbots. 

\begin{figure}[htbp]
    \centering
    \includegraphics[width=7.7cm]{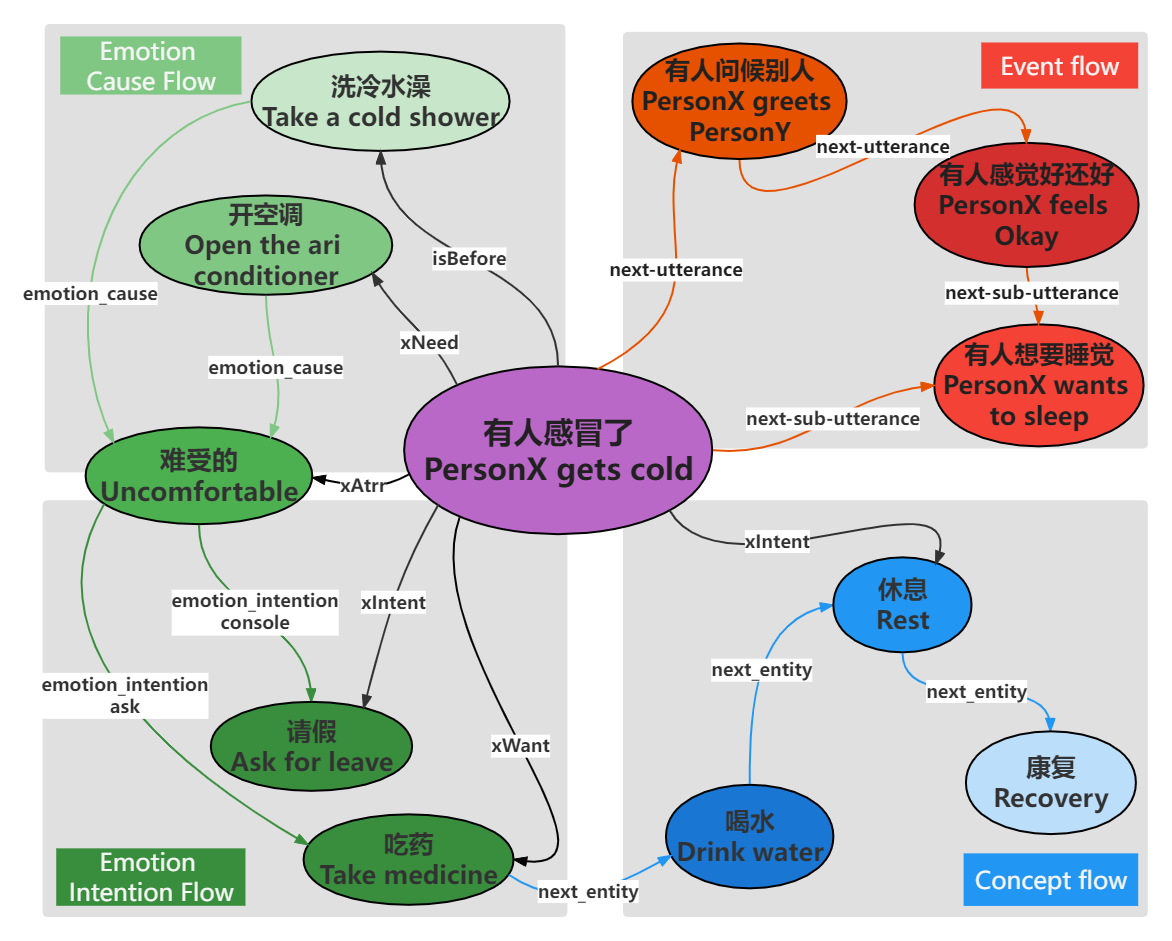}
    \caption{A tiny subset of C$^3$KG, with four unique types of dialog flow relations.}
    \label{fig:graph}
\end{figure}

In spite of the potential, it has two major difficulties. For instance, when a friend in \emph{distress} tells us that he recently adopted a cat, we humans will easily suspect that he might has allergies to the cat. However, such reasoning is difficult for chatbots. Given the event-relation pair \{\emph{PersonX adopts a cat}, \verb|xEffect|, \_\_\_\}, ATOMIC contains multiple tails like \{\emph{finds out he has allergies}\} and the tail \{\emph{becomes less lonely}\}. To this end, the first difficulty comes from the existence of multiple tails, which will confuse the chatbots when inferring the cause behind the negative emotion. Secondly, the knowledge tuples in ATOMIC are isolated. It is thus more difficult for the chatbots to reason which tail(s) of knowledge should be used to produce coherent responses. For example, if the tuple \{\emph{PersonX adopts a cat}, \verb|isAfter|, \emph{finds a cat at the animal shelter}\} is detected from the dialogue history, then the tuple \{\emph{PersonX adopts a cat}, \verb|xNeed|, \emph{go to an animal rescue center}\} should not be considered anymore for future conversations. We argue that these issues hamper the application of ATOMIC to multi-turn dialogue modeling where the conversational agents need not only know the current state but also plan the future \textbf{dialog flow}. 

To remedy these issues, we define 4 novel dialog flow relations, i.e., event flow, concept flow, emotion-cause flow, emotion-intent flow, as depicted in Figure~\ref{fig:graph}. To build up the relations, we collect a large-scale multi-turn conversations in everyday scenarios, and manually annotate the conversations with emotional information. Based on the annotations, we are able to extract conversation-related events in ATOMIC and connect them using different dialog flows. In this way, we augment ATOMIC with conversation-specific knowledge, which facilitates chatbots to pick out useful commmonsense knowledge, and relieves their confusion on noisy knowledge that are incoherent with dialog flows. We believe our graph is favorable for commonsense conversation modeling. 

To highlight: (1) We curate a new Chinese corpus, containing multi-turn human-written conversations on dailylife topics and rich, high-quality annotations on the level of sub-utterance; (2) We create and will release the first large-scale \textbf{C}hinese \textbf{c}ommonsense \textbf{c}onversation knowledge graph, \textbf{C}$^{3}$\textbf{KG}, which contain 4 types of unique dialog-flow edges to store the distilled conversation knowledge from the multi-turn conversation corpus; (3) We devise a graph-conversation matching approach, and benchmark 2 typical tasks grounded on commonsense conversation graph.

\section{Related Work}
\subsection{Commonsense Knowledge Bases}
ConceptNet~\cite{Speer2017ConceptNet5A} is a popular commonsense knowledge base (CKB), which has a Chinese version with a relatively small set of knowledge~\cite{Kuo2009CommunitybasedGD}. Another large-scale CKB is TransOMCS~\cite{Zhang2020TransOMCSFL}, which is built automatically by converting syntactic parses of Web sentences into structured knowledge. However, the majority of relations in existing CKBs are taxonomic relations such as \verb|isA| and \verb|Synonym|~\cite{Davis2015CommonsenseRA}, which inevitably limits their capabilities. Differently, we rely on mental CBK ATOMIC~\cite{atomic}, translate ATOMIC into Chinese and build dialog flow relations on it, with the aim of facilitating Chinese conversational systems.  

To construct these CKBs, ATOMIC and ConceptNet rely on crowd-sourcing by which annotators add tail knowledge to a given entity or event based on their own commonsense. To improve efficiency, \citet{bosselut2019comet} propose COMET, a pre-trained language model which is able to generate diverse tail knowledge given any new event. This automates the collection procedure and results in a scaling of commonsense knowledge. Nevertheless, \citet{Zhang2020TransOMCSFL} argues that COMET still suffers from overfitting problem and tends to produce high-frequent and repetitive  knowledge. To address, they develop DISCOS~\cite{fang2021discos} that learn the extracting patterns from existing CKBs and automatically distill commonsense knowledge from the AESR knowledge graph~\cite{zhang2020aser}.

\subsection{Connecting Knowledge and Conversation}

One line of work attempts to extract structured knowledge from conversations. These works detect named entities from each utterance in conversational datasets~\cite{xu2020conversational,zou2021thinking,ghosal2021cider} and build up the relationship based on their sequential order and Pointwise Mutual Information (PMI)~\cite{church1990word}. There also exist some works that adopt automatic extraction tools, such as OpenIE, to construct conversational knowledge bases of certain domains~\cite{ahmad2020active}. Although plausible, these knowledge graphs are built on the granularities of word or phrase, which makes them hard to match the overall semantics of dialogue sentences. In this paper, we build a Chinese commonsense conversation knowledge graph based on both multi-turn conversational corpus and event-centered knowledge base. At the same time, we propose to use Sentence-BERT~\cite{reimers2019sentence}, a transformer-based semantic similarity model, to construct dialog flow edges in our knowledge graph. 

There is also another line of growing interests in incorporating commonsense knowledge into conversation modeling. Both \citet{zhou2018commonsense} and \citet{zhang2019grounded} introduce knowledge triplets from ConceptNet~\cite{speer2017conceptnet} into open-domain response generation. Recently, \citet{mkedg} and \citet{care} exploit ConceptNet to enhance emotion reasoning for response generation, and others design graph reasoning methods to plan the topic transition in the responses~\cite{Moon2019OpenDialKGEC,Tang2019TargetGuidedOC,Xu2020KnowledgeGG,Li2021GraphStructuredCU}. One distinct work is \citet{ghosal2020cosmic}, which utilizes ATOMIC~\cite{hwang2020comet} in emotional dialogue modeling for emotion identification. In this paper, we connect the heads and tails in ATOMIC according to four types of dialog flows. Because the resulted graph C$^3$KG contains both social knowledge from ATOMIC and dialogue knowledge from our corpus, it is thus more suitable for empathetic conversation modeling.

\section{A Scenario-based Multi-turn Conversation Corpus}
\label{sec:corpus}
Our aim is to extract common dialog flow information from real conversations. In this way, it is crucial to ensure the quality of the conversation corpus and the reliability of the extraction method. In the following, we firstly introduce the conversation corpus \textbf{CConv} we depend on. 

Instead of using the noisy Internet data, we collect a multi-turn human-written Chinese conversation corpus based on crowdsourcing. Initially, 100 workers are hired, and they are randomly paired to talk in text under a given scenario. Each scenario is one sentence describing the suggested conversation context which often involves certain everyday events. Besides, the workers are also required to follow certain rules like ``each utterance should longer than 6 Chinese characters'', which are critical to help ensure the quality of the collected conversation. At the beginning of the crowdsourcing, we check each collected conversation and re-train the workers. To ensure the quality, we keep only 62 well-trained workers and let them finish our task. Note that the workers are paid with 1 CNY per utterance (nearly 0.2 dollar per utterance). Finally, we obtain 32k sessions of high-quality two-party conversations (650k utterances in total) on 200 scenarios of 15 daily topics. 

To facilitate future research, we then hire another 3 well-trained assistants to manually annotate the conversations with fine-grained emotional labels including speaker's emotion type, emotion cause, and response intention type. Following~\citet{Rashkin2019TowardsEO}, we define emotion type with 5 general classes \{joy, angry, sad, surprising, other\}. Emotion cause span is a continuous text spans implying the reason of certain emotion~\cite{li2021towards}. Response intention type is essential for building empathetic chatbots, and we define 6 commonly-adopted intent classes of \{ask, advise, describe, opinion, console, other\} following~\citet{Welivita2020ATO}. A snippet of a conversation example is given in Figure~\ref{fig:pipeline}. In Appendix, we present more information of the constructed corpus.

By utilizing the annotations, we are able to distill dialogue knowledge to enhance the conversation graph and graph-grounded conversation modeling.

\section{Overview and Processing of ATOMIC}
Because our conversation corpus is Chinese, we want to build a Chinese conversation knowledge graph. It is well known that to build a knowledge graph from scratch is laborious and time-consuming. 
Instead, we base on ATOMIC and design a pipeline method to translate it into Chinese, meanwhile ensuring the resulted knowledge graph is reliable and suitable for conversation grounding.

\subsection{Brief Introduction of ATOMIC}
We firstly give a brief description of ATOMIC~\cite{atomic}. ATOMIC organizes commonsense knowledge in the form of triplet <head, relation, tail>, where head often describes a daily event.

There are two unique properties making ATOMIC suitable and attractive for building empathic chatbots. Firstly, ATOMIC collects knowledge about how people will react to a given event. This kind of knowledge is related to people's mental states, which is beneficial for understanding implicit emotions. For example, given a head event \emph{PersonX makes PersonY’s coffee}, ATOMIC contains knowledge that PersonY will be \emph{grateful} along the relation \verb|oReact|. 
Secondly, ATOMIC organizes knowledge using several inferential relations and naturally supports \emph{if-then} reasoning, which is crucial generating coherent responses. Totally, there are 9 relations defined in ATOMIC. The details can be found in Appendix. 

In the terms of translating ATOMIC to Chinese, we apply \textbf{Regular Replacement} and \textbf{Joint Translation} method to improve the quality of translation. We give more details of our translation methods in the Appendix. we denote the translated ATOMIC as \textbf{ATOMIC-zh}.

\begin{figure}[!t]
    \centering
    \includegraphics[width=7.7cm]{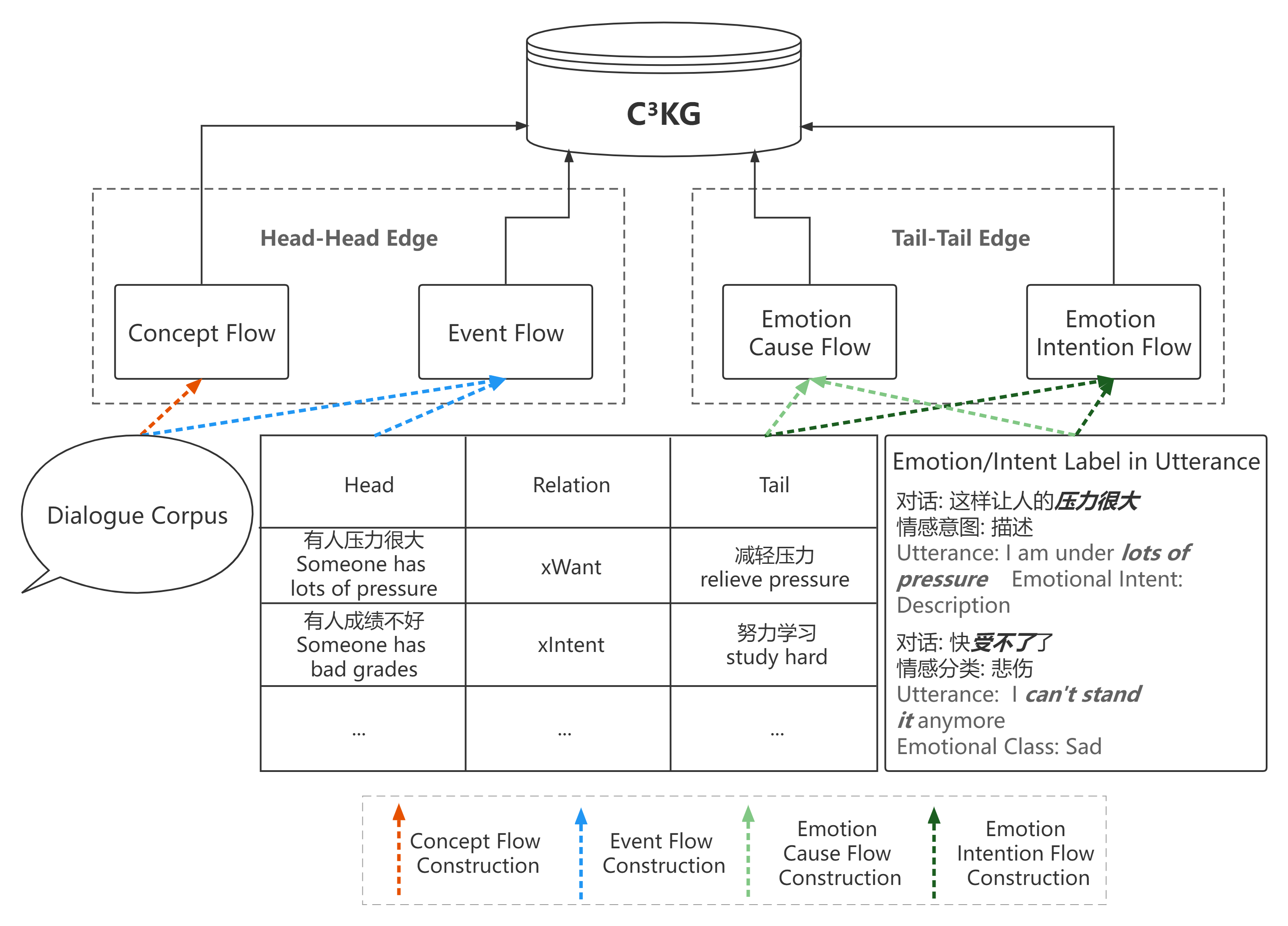}
    \caption{Construction Process of C$^3$KG.}
    \label{fig:pipeline}
\end{figure}

\section{Conversation Knowledge Graph Construction}
\subsection{Overview of C$^3$KG}
To supply dialog flow information for commonsense reasoning, we create a \textbf{C}hinese \textbf{C}ommonsense \textbf{C}onversation \textbf{K}nowledge \textbf{G}raph, \textbf{C}$^3$\textbf{KG}, whose statistics are summarized in below.

\begin{table}
\centering
\begin{tabular}{|l|c|c|}
\hline
\multirow{5}{*}{\#Relations} & ATOMIC Relations & 636,636\\\cline{2-3}
& Event Flows   &  571,196\\
& Concept Flows  & 77,587\\\cline{2-3}
& Emotion-Cause Flows & 269\\
& Emotion-Intent Flows & 553\\\hline
\#Triplets & \multicolumn{2}{|c|}{1,286,241} \\
\hline
\end{tabular}
\caption{Statistics of C$^3$KG.}
\label{tab:graph_stats}      
\end{table}

We then introduce our method of constructing a conversational knowledge graph based on ATOMIC-zh and our multi-turn conversation corpus. In general, we extract events from each conversations and match with the head in ATOMIC-zh. The core is how to build new dialog flow relations, which is depicted in Figure~\ref{fig:pipeline}, and will be detailed present in the following section.

\subsection{Event Extraction}
\label{Event Extraction}

Knowledge in ATOMIC-zh is event-based and most of them are declarative sentences with some entities omitted. However, utterances in the open-domain dialogue dataset contain a lot of colloquial expressions and sub-sentences with more complex structures. To address, we develop a dependency parsing-based event detection pipeline to extract salient events in each utterance. 
The overview of our algorithm is described in Algorithm~\ref{tab:algorithm}.

\noindent\textbf{Pre-processing}. We first split each utterance with punctuation, and operate on the level of sub-utterances. 
To reduce noise, we then filter short sub-utterances with transitive and dumb semantics like \begin{CJK*}{UTF8}{gkai}``好的''\end{CJK*} (OK),  \begin{CJK*}{UTF8}{gkai}``就是这样''\end{CJK*} (That's it). After that, we perform Dependency Syntactic Parsing and POS tagging using ltp4\footnote{\url{https://github.com/HIT-SCIR/ltp}}, and extract event mentions based on two kinds of structural patterns, verb-driven and adjective-driven clauses.

\begin{algorithm}[t]
\caption{Event Extraction from Utterance} 
\label{alg:A}  
\hspace*{0.02in}{\bf Input:}
An utterance $U$\\
\hspace*{0.02in}{\bf Output:}
A set of event mentions $M$
\begin{algorithmic}[1] 
\STATE Split $U$ with punctuation, and get a series of sub-utterance $SU$, filter $SU$ based on length
\FOR{each $su \in SU$} 
    \STATE Obtain the dependency tree $dep$ and POS tagging result $pos$ of $su$ 
    \STATE Find the $had$ node which connects directly to the $ROOT$ node in the dependency tree
    \IF{POS tag of the $had$ node $\in$ [$v$, $a$]}
        \STATE Append $had$ to $HAD$
    \ENDIF
    \IF{The number of verbs connected directly to $had$ more than $1$}
        \STATE Recursively search verbs in the sub-tree of $had$ and replace $had$ in $HAD$ with the founded verbs
    \ENDIF
    \FOR{$had \in HAD$}
        \IF{POS of node $had$ is $v$}
        
            \STATE Keep words in $su$ that appear after $had$ and words connect directly to $had$ and relation is `ADV', connect them and append to $M$  
        \ELSE
            \STATE Remain words in $su$ that connect directly to $had$ and relation is `SBV', connect them and append to $M$
        \ENDIF
    \ENDFOR
\ENDFOR
\STATE Return $M$
\end{algorithmic}
\label{tab:algorithm}
\end{algorithm}

\noindent\textbf{Verb-driven}. Verb-driven clauses have a verb connecting to the root node in the dependency tree. After filtering some noisy words, we obtain verb-driven event mentions. For example, we extract the mention \begin{CJK*}{UTF8}{gkai}``催促提供物资的商家''\end{CJK*} (urged the merchants who provide supplies) from utterance \begin{CJK*}{UTF8}{gkai}``我和上司已经在催促提供物资的商家了''\end{CJK*} (My boss and I have already urged the merchants who provide supplies). In this utterance, we filter subject of utterance\begin{CJK*}{UTF8}{gkai}``我和上司''\end{CJK*} (My boss and I), adverbial\begin{CJK*}{UTF8}{gkai}``已经''\end{CJK*} (have already) and modal particle\begin{CJK*}{UTF8}{gkai}``了''\end{CJK*} (yet) at the end of the utterance.

\noindent\textbf{Adjective-driven}. Besides, adjective-driven clauses often have meaningful entities in sub-utterances. Similarly, we extract adjective-driven event mentions based on the adjective-driven clauses by keeping the modifier of its key adjective and filtering out other words. For example, we extract the mention \begin{CJK*}{UTF8}{gkai}``学习节奏快''\end{CJK*} (The pace of learning is fast) from the utterance \begin{CJK*}{UTF8}{gkai}``但学习节奏也太快了吧''\end{CJK*} (But the pace of learning is too fast). In this utterance, we filter the initial conjunction \begin{CJK*}{UTF8}{gkai}``但是''\end{CJK*} (but), adverbial \begin{CJK*}{UTF8}{gkai}``也''\end{CJK*} (no meaning) and \begin{CJK*}{UTF8}{gkai}``太''\end{CJK*} (too) and modal particle \begin{CJK*}{UTF8}{gkai}``了''\end{CJK*} (yet) and \begin{CJK*}{UTF8}{gkai}``吧''\end{CJK*} (no meaning) at the end of the utterance. 

\noindent\textbf{Recursive Applying}. The resulted event mentions may still contain multiple verbs and several semantic units. In this case, we apply a secondary decomposition. 
For example, we will split the event mention \begin{CJK*}{UTF8}{gkai}``以为进了大学就可以放松放松''\end{CJK*} (could relax after entering university) into two events \begin{CJK*}{UTF8}{gkai}``进了大学''\end{CJK*} (entering university) and \begin{CJK*}{UTF8}{gkai}``就可以放松放松''\end{CJK*} (could relax). To do so, we count the number of verbs connected to the root word in the mention as well as the depth of the sub-trees led by those verbs. Based on the results, we determine whether the mention needs a secondary decomposition using a threshold. If needed, we recursively search verbs in the original dependency tree and replace the key verb with the verbs we found.

\subsection{Event Linking as Matching}
\label{Matching and construction}

\begin{figure}[htbp]
    \centering
    \includegraphics[width=7.7cm]{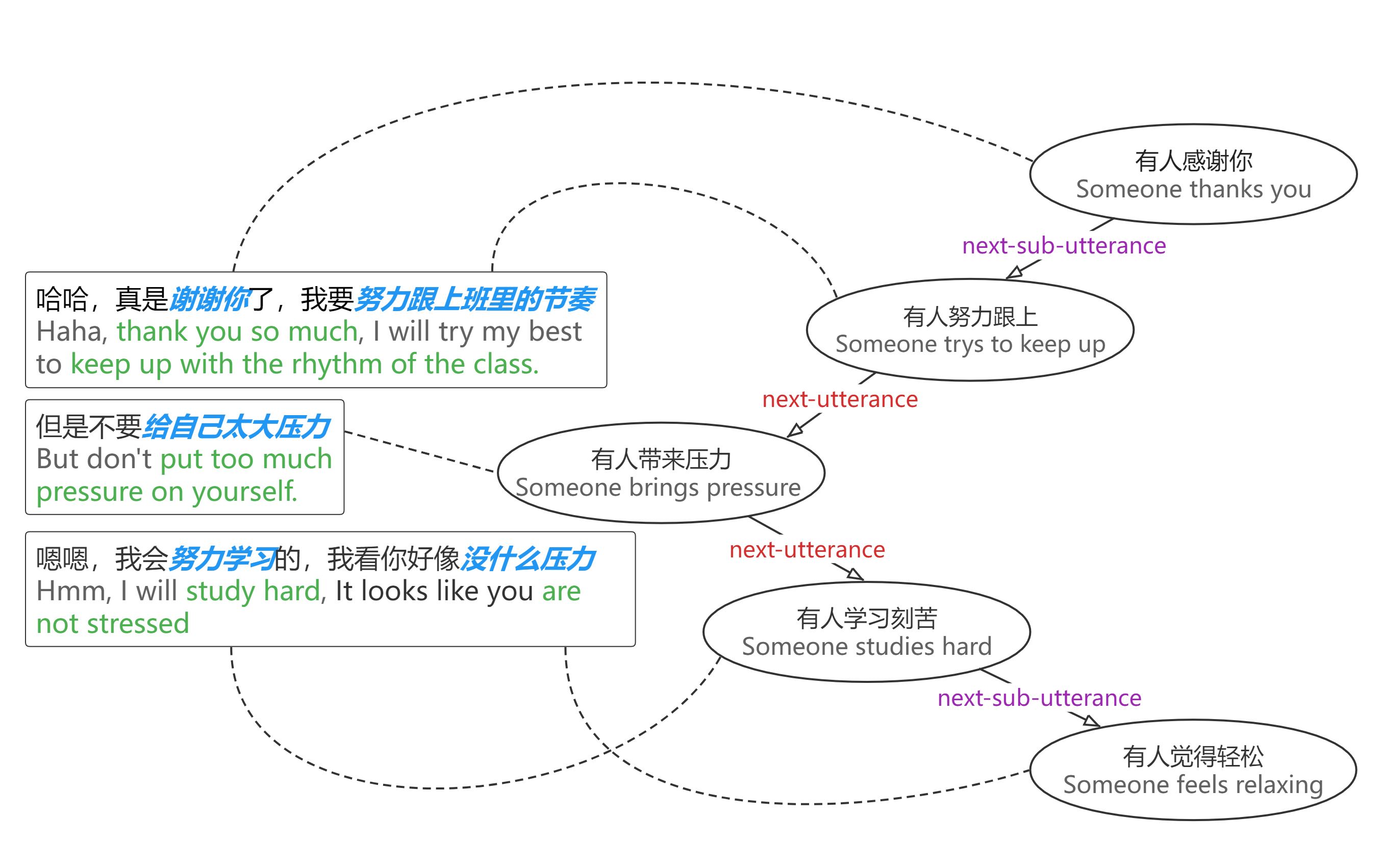}
    \caption{An Example of Head-Head Edge Construction for Event Flows.}
    \label{fig:matching}
\end{figure}

In order to discover common dialog flows among the knowledge base, the event mentions in the conversations are then linked to ATOMIC heads using matching techniques. 

Typically, we adopt Sentence-BERT, a powerful semantic matching model, which is based on Siamese and Triplet Network and pre-trained on sentence pairs in different relationships~\cite{reimers-gurevych-2019-sentence}. It encodes two given sentences separately and calculates the similarity between their representations, and thus performing efficiently in large-scale many-to-many matching.

To enhance the matching performance, we finetune Sentence-BERT on our corpus. Specifically, we randomly select 8,000 <$m$, $h$> mention-head pairs matched by pre-trained Sentence-BERT, and manually label a matching score in \{0,1\} for fine-tuning. Note the reason why we adopt discrete 0,1 instead of continuous $[0,1]$ scores is that using the former effectively mitigates the domain gap. It will induce the matching model to label 0 for those <$m$, $h$> share similar characters in surface but different meanings in semantics. After fine-tuning, we calculate the cosine similarity scores and choose the head with the highest score as the matching result given an event mention.

\subsection{Edge Construction}
Now we have 32k sessions of multi-turn conversations and link their event mentions to ATOMIC heads. The remaining is how to utilize them and build commonsense conversation knowledge graph. In this work, we propose three kinds of edges to reflect different types of dialog flows.

\subsubsection{Head-Head Edge Construction}
\noindent\textbf{Event Flow}. Naturally, a dialogue is hierarchical in that it consists of a sequence of utterances produced by two interlocutors, where each utterance is composed of one or several sub-utterances. If two event mentions are detected together within in a conversation, the co-occurrence can be regarded as a dialog flow example. Following the flow, it is then intuitive to connect the ATOMIC heads linked by the mentions, as illustrated in Figure~\ref{fig:matching}. By connecting intra-utterance and inter- utterance mentions, we acquire the event flows of \verb|next-sub-utterance| and \verb|next-utterance|.

\noindent\textbf{Concept Flow}. ATOMIC also has entity-level heads in addition to the phrase-level events. To utilize them, we perform entity linking by detecting word entities with POS tag belonging to \{verb, noun, adjective\} in the original conversations, and match them with the entity-level ATOMIC heads to construct concept flow edges similarly. These concept flows are helpful for planning and transiting the contents in topic-aware conversation~\cite{Yao2018ChatMI,Moon2019OpenDialKGEC,Xu2020ConversationalGG,zou-etal-2021-thinking}.

Because we are interested in the most common dialog flows, we only keep those highly-frequent connections, and create a head-to-head dialog flow between the ATOMIC head entities and events.

\subsubsection{Tail-Tail Edge Construction}
Besides, we also consider another essential type of dialog flow, i.e., emotion-based empathy flow. 

In this paper, we utilize the emotional labels on our corpus (in Section~\ref{sec:corpus}) to construct two kinds of emotion-based edges connecting tails in our knowledge graph. Intuitively, \verb|emotion-cause| dialog flow reflects the reasons for a specific emotion, which is useful for fine-grained emotion understanding. And \verb|emotion-intent| empathy flow indicates what response intentions are proper to use when the other one is in a specific emotion, which is critical for response empathy.

\noindent\textbf{Pre-processing}. To construct emotion-based edges, we category the tails into 3 classes according to their connecting relations, as listed in Table~\ref{tab:emotion_relation}. The first class of tails are linked by relations $xAttr$ or $xReact$, which reflects people's psychological reaction towards a certain event (head). For instance, \{\emph{PersonX runs out of steam}, \verb|xAttr|, \emph{tired}\} indicates that someone is lacking energy. We denote the first class as Tail$_{emotion}$. The second 
class Tail$_{before}$ states the events commonly happen before the heads, e.g., \{\emph{PersonX runs out of steam}, \verb|isAfter|, \emph{PersonX exercises in gym}\}. On the contrary, the last class Tail$_{after}$ contain the events following the head events like \{\emph{PersonX runs out of steam}, \verb|xWant|, \emph{to get some energy}\}. 

By analyzing these relations and tails, we find heuristics to build emotion-based dialog flows. By connecting the head and tails in class Tail$_{emotion}$, we are able to create causal emotional inference like \{\emph{PersonX exercises in gym}, \verb|emotion-cause|, \emph{tired}\}. Through cross linking the tails in class Tail$_{emotion}$ and Tail$_{after}$, we are able to develop the inferential edges like \{\emph{tired}, \verb|emotion-intent|, \emph{to get some energy}\}.

\begin{table}
\centering
\begin{tabular}{|l|c|}
\hline
Tail$_{emotion}$ &\makecell[l]{xAttr,xReact}\\
\hline
Tail$_{before}$ &\makecell[l]{isAfter, xNeed}\\
\hline
Tail$_{after}$ &\makecell[l]{isBefore, xWant, xIntent, \\xEffect, oEffect} \\
\hline
\end{tabular}
\caption{Relation Categories For Emotion-based Edge Construction.}
\label{tab:emotion_relation}       
\end{table}

\noindent\textbf{Filtering}. Based on the heuristics, we apply SentiLARE\footnote{https://github.com/thu-coai/SentiLARE} to  match each tail in class Tail$_{emotion}$ to one of 4 emotion labels defined in our dataset, i.e., \{joy, sad, angry, others\}. For label 'surprising' (which is not contained in the labels of SentiLARE), we use Sentence-BERT\footnote{This model is not fine-tuned on our dataset.} and set a threshold of 0.7 to label 'surprising' in the tails whose label is 'others' according to SentiLARE. The tails sharing the same emotion class with the original utterance are kept to build emotion-based dialog flows. 

\noindent\textbf{Emotion Cause Flow}. Then, we apply keyword-based exact matching between the tails in Tail$_{before}$ with dialogue context. For Tail$_{before}$, if there is an keyword exactly matched with some keywords in the previous utterances, we create an $emotion-cause$ edge flowed from the tail of Tail$_{before}$ to those filtered tails in Tail$_{emotion}$, indicating that the event of Tail$_{before}$ may cause person to feel the emotion of the tail in Tail$_{emotion}$.

\begin{figure}[htbp]
    \centering
    \includegraphics[width=7.7cm]{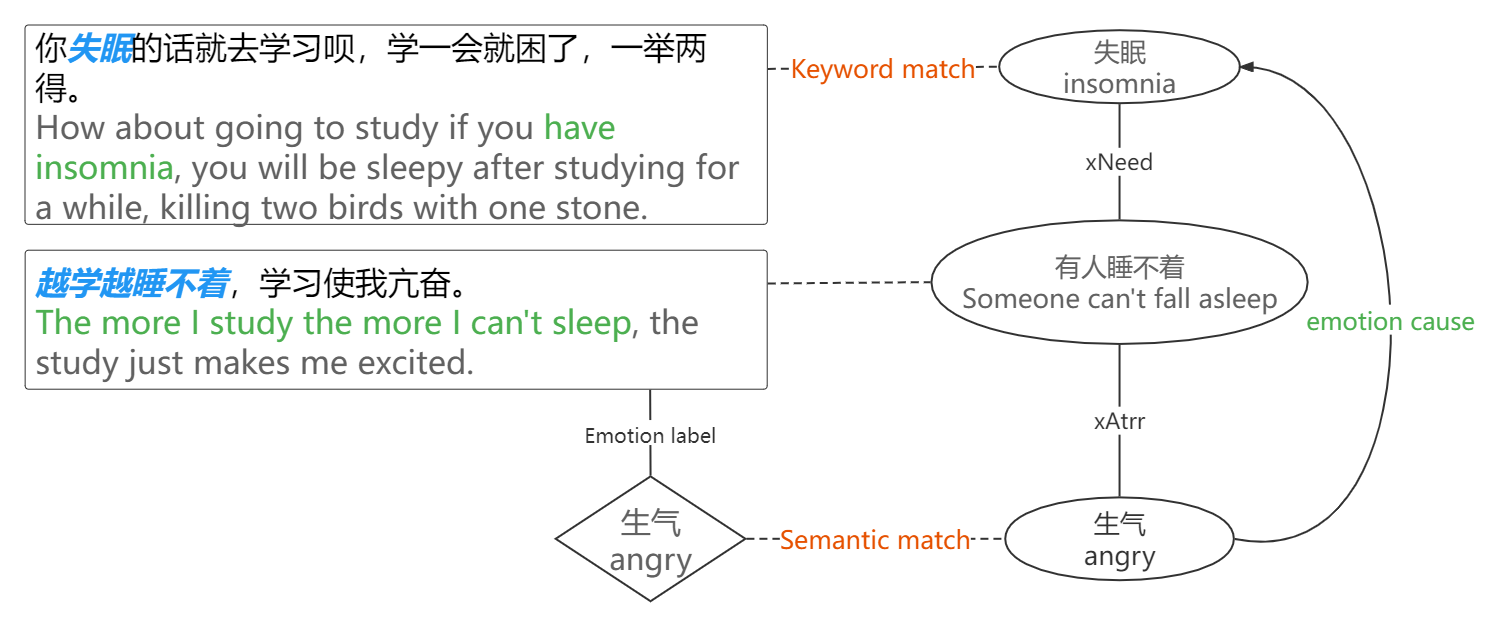}
    \caption{An Example of Tail-Tail Edge Construction for Emotional Cause Flows.}
    \label{fig:emotion cause}
\end{figure}

Figure~\ref{fig:emotion cause} depicts the process of constructing the labeled \verb|emotion-cause| edge. Firstly, we match the tail \emph{angry} in Tail$_{emotion}$ to the utterance emotion label "angry". Then, we detect that the tail \emph{insomnia} in Tail$_{before}$ shows up in the previous utterance. So we build a \verb|emotion_cause| edge from the tail \emph{angry} to tail \emph{insomnia}. This kind of tail-tail \verb|emotion_cause| flows is supportive for chatbots to have a better understanding of users' emotional mood by reasoning its cause.

\noindent\textbf{Emotion Intent Flow}. For tails in class Tail$_{after}$, we create an \verb|emotion_intent| flow from those filtered tails in Tail$_{emotion}$ to the tails in Tail$_{after}$. Notably, we also assign one of five intent labels to each \verb|emotion_intent| edge, i.e., \{ask, advise, describe, opinion, console\} (Section~\ref{sec:corpus}).  

\begin{figure}[htbp]
    \centering
    \includegraphics[width=7.7cm]{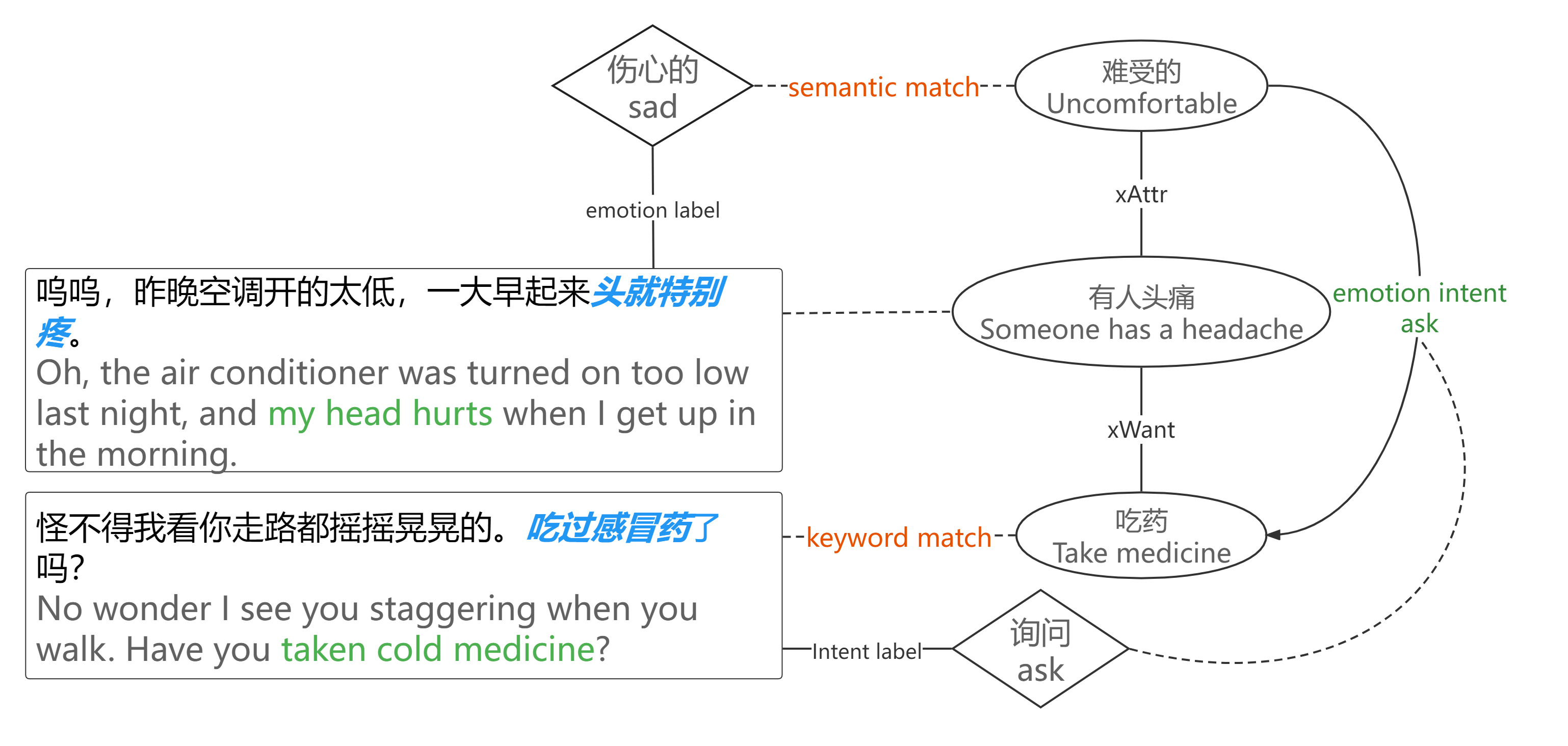}
    \caption{An Example of Tail-Tail Edge Construction for Emotional Intent Flows.}
    \label{fig:emotion intention}
\end{figure}

Figure~\ref{fig:emotion intention} depicts the process of constructing the labeled \verb|emotion-intent| edge. We start by matching the tail \emph{Uncomfortable} in Tail$_{emotion}$ to the utterance emotion label "sad". Then, we detect that the tail \emph{Take medicine} in Tail$_{after}$ shows up in the next utterance. As such, we build a \verb|emotion_intent| edge from the tail \emph{Uncomfortable} to tail \emph{Take medicine}, and add the intent label of the second utterance ``ask'' on to the edge. This kind of tail-tail \verb|emotion_intent| flows is supportive for chatbots to choose proper response strategy under a certain situation. 

\noindent\textbf{Expertise Label}. Considering that both emotion and intent within each utterance is latent and subtle, it is very hard to make the emotion flow results of automatically extraction behave well in the terms of number. In that case, we also hire 2 expertise with rich experience in psychology, and hire them to label both emotion cause and intent in high-frequency scenarios for emotion expression, like sleeplessness and academic pressure.

For expertise convenience, we also build an interactive annotation tool for more easily annotating and exploring in our C$^3$KG. The system integrates functions like revising and adding tails, which would be a good supplement and cleaning tool for our C$^3$KG. There are more details of our tool in the Appendix.

\section{Evaluation}
\subsection{Matching Evaluation}
\noindent\textbf{Manual Assessment}. We randomly choose 100 utterances to evaluate our event extraction (Section~\ref{Event Extraction}) and matching methods (Section~\ref{Matching and construction}). We denote our proposed method as $Parsing$. To compare with it, we use another two methods to process utterances: $POS$ employs POS tagging-based templates to extract events, and $Simple$ only splits and filters utterances according to punctuation before matching. We report matching results using both Sentence-BERT and Sentence-BERT-finetune.

In Table~\ref{tab:matching evaluation}, Similarity stands for the averaged matching degree, and Number for the average number of matched ATOMIC heads of the chosen utterances, which can be seen as an indicator for matching recall. Although the three methods have similar average similarity without finetuning, our Parsing method gets an obvious similarity improvement after finetuning as compared with Simple and POS without loss of knowledge recall, which is also significantly better than POS-based method. 

\begin{table}[htbp]\small%
  \centering
  \begin{threeparttable}  
    \begin{tabular}{ccccc}  
    \toprule  
    \multirow{2.5}{*}{Method}&  
    \multicolumn{2}{c}{SBERT}&\multicolumn{2}{c}{SBERT-finetune}\cr  
    \cmidrule(lr){2-3} \cmidrule(lr){4-5}  
    &Similarity&Number&Similarity&Number\cr  
    \midrule  
    Simple&51.3$\%$&{\bf 1.57}&53.2$\%$&{\bf 1.57}\cr  
    POS&{\bf 51.4$\%$}&0.75&54.1$\%$&0.75\cr  
    Parsing&51.3$\%$&1.53&{\bf 55.3$\%$}&1.53\cr  
    \bottomrule  
    \end{tabular}  
  \caption{Comparison of Matching Approaches.}
    \label{tab:matching evaluation}
    \end{threeparttable}  
\end{table}

\noindent\textbf{Scenario Graph Visualization}. We also build up scenario graphs based on matching results and the scenario descriptions. By visualizing the matched result for each topic of scenarios, we are able to better understand the matching quality.

\begin{figure}[!t]
    \centering
    \includegraphics[width=7.7cm]{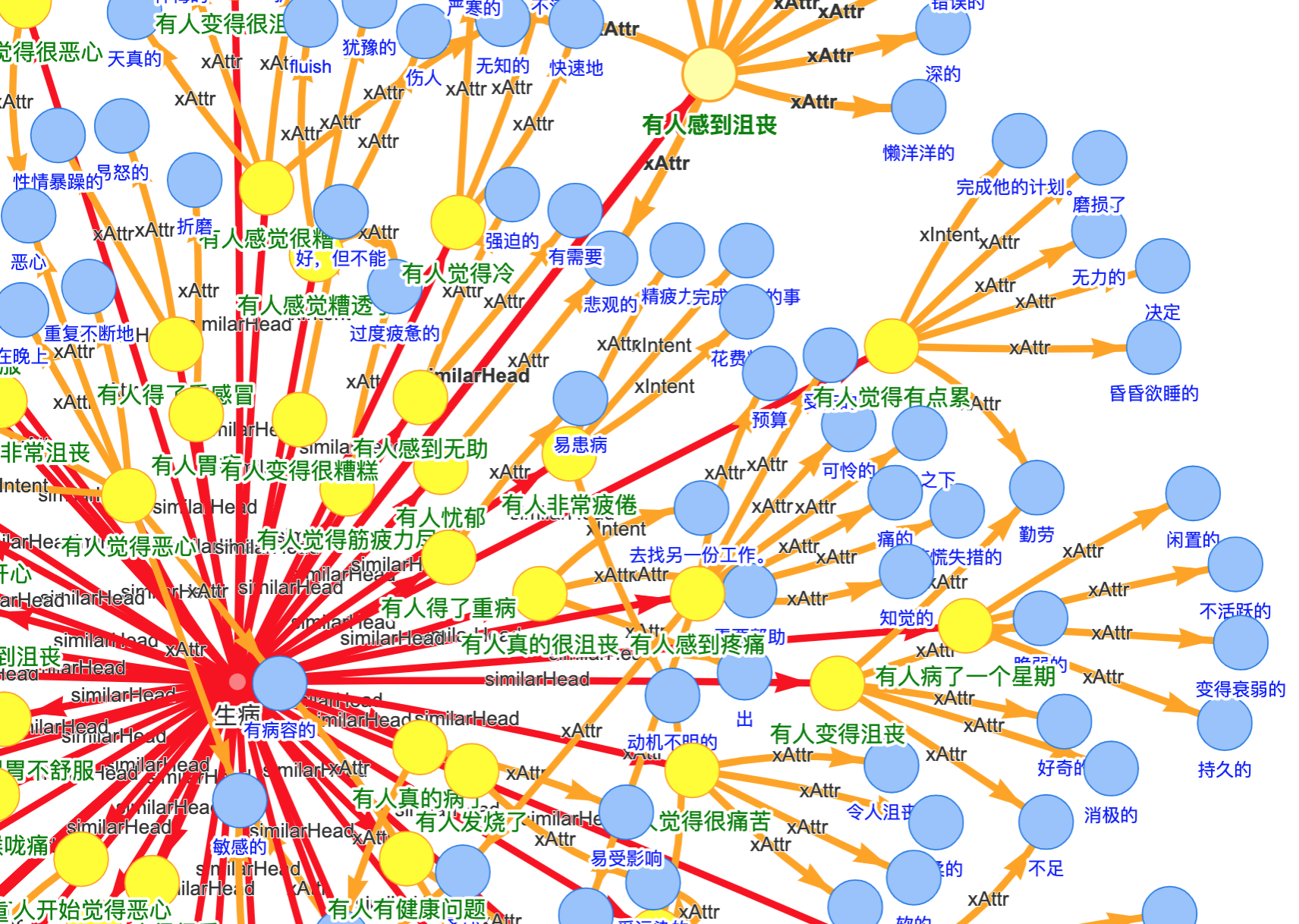}
    \caption{Scenario Graph of ``Sickness''.}
    \label{fig:illness}
\end{figure}

Specifically, we use sub-sentence to match heads in ATOMIC-zh, and use the top 0.5\% heads we match in each scenario to build scenario-based graphs. Each of them can be seen as a sampled sub-graph from ATOMIC-zh, with higher topic coherence with its scenario. After annotation, the matching accuracy based on 3 annotators reaches 0.71, which indicate a fair quality of scenario graph. To depict, we visualize a snippet of the scenario graph ``sickness'' in Figure~\ref{fig:illness}. Please kindly note that for clarity, we only visualize a small set of relation and tails in Figure~\ref{fig:illness}. In fact, every scenario graphs contain the full set of C$^3$KG relations. For more scenario graphs, please check Appendix.

\subsection{Graph Evaluation}

\noindent\textbf{Node Evaluation}.
\label{sec:node_evaluation}
Since our C$^3$KG is built upon the translated ATOMIC-zh. We firstly evaluate the quality of our graph in terms of translation accuracy. In specific, we randomly sample 200 triplets from C$^3$KG, and ask annotators to label each Chinese triplet in terms of fluency and logic correctness with \{0,1\} scores. To validate our joint translation method, we also compare with the results using separate translation. 

\begin{table}[!t]
\centering
\small
\begin{tabular}{ccc}
\hline
Method & Fluency & Logic      \\ 
\hline
Separate translation & 0.825 & 0.71                    \\
Joint translation&{\bf 0.92} &{\bf 0.88}                  \\
\hline
\end{tabular}
\caption{Evaluation of Translation Quality.}
\label{translation quality}
\end{table}

As shown in Table~\ref{translation quality}, the significant increases on both Fluency and Logic aspects clearly demonstrate the superiority of joint translation method. In terms of logical coherence, we find many sample cases are labeled with $0$ logical score due to the incompleteness of their heads, which somehow confuses the semantics and obstacles logical connection to the tails. For example, \{\emph{{\begin{CJK*}{UTF8}{gkai}有人把他父亲\end{CJK*}}}, \verb|xAttr|, \emph{{\begin{CJK*}{UTF8}{gkai}告密者\end{CJK*}}}\} (\{\emph{PersonX gets PersonX's father}, \verb|xAttr|, \emph{a tattletale}\}) seems ridiculous. However, if we add \emph{{\begin{CJK*}{UTF8}{gkai}叫来\end{CJK*}}} (\emph{came}) in the end of the heads, then we could imagine a scenario where a child threatens another child by summoning parents.
Nonetheless, such seemingly illogical knowledge might still be informative for downstream tasks with fuzzy matching techniques. Hence, we retain this kind of incomplete heads.

\noindent\textbf{Edge Evaluation}.
At the heart of C$^3$KG is the novel dialog flow relations we develop in this work. To validate the quality and robustness of these relations, we utilize another open-domain multi-turn Chinese dialogue dataset, MOD~\cite{mod}\footnote{\url{https://github.com/lizekang/DSTC10-MOD}}. In specific, we extract event mentions from MOD utterances and match them to our graph using the methods as in Section~\ref{Event Extraction}. Then we evaluate the connectivity and average distance of the matched results, w.r.t. both \verb|next_utterance| and \verb|next_sub_utterance| relations. This aims to assess the aggregation degree of related content in our knowledge graph.

\begin{table}[htbp]
\small%
\centering
\begin{tabular}{@{}ccccc@{}}
\hline
\multirow{2.5}{*}{KG} &
\multicolumn{2}{c}{next\_utterance} & \multicolumn{2}{c}{next\_sub\_utterance} \\
\cmidrule{2-5} & Con. & AVG\_Dist. & Con. & AVG\_Dist. \\
\hline
C$^3$KG &{\bf 96.68$\%$}& {\bf1.86}& {\bf 96.51$\%$}& {\bf 1.90} \\
ATOMIC-zh &6.90$\%$& 7.52& 5.21$\%$& 10.81 \\
\hline
\end{tabular}
\caption{Edge Evaluation Result on MOD dataset.}
\label{edge evaluation}
\end{table}

Table~\ref{edge evaluation} shows our edge evaluation result on MOD. For comparison, we add the test result of ATOMIC-zh, considering their similarity in size. The comparing result shows the effectiveness of our event flow, which leads the matching of context within a dialogue has higher connectivity and shorter distance.

\begin{table}[!t]
\centering
\small
\begin{tabular}{|c|c|c|}
\hline
{Method} & Emotion (acc.) & Intent (acc.) \\
\hline
Base            & 90.7$\%$  & 65.3$\%$    \\
Knowledge       & {\bf93.6$\%$}    & {\bf71.3$\%$}   \\
History      & 90.5$\%$     & 64.7$\%$     \\
Knowledge+History  &  91.2$\%$  &  67.4$\%$     \\ 
\hline
\end{tabular}
\caption{Baselines for Graph-grounded Tasks.}
\label{proposed tasks}
\end{table}

\section{Proposed Tasks}
To show the potential, we propose two graph-grounded conversational tasks, i.e., emotion classification and intent prediction, and train benchmark models using our labeled corpus CConv.

\noindent\textbf{Task 1: Emotion Classification} requires to produce an emotion label conditions on the conversations. Following common practice, we choose the BERT model, and sample the \verb|xAttr|, \verb|xReact| tails from our matching head as extra input.

\noindent\textbf{Task 2: Intent Prediction} requires to predict a proper type of response intent for the conversations. We choose BERT model, and sample the \verb|oReact|, \verb|oEffect| tails from our matching heads. As simple baselines, we introduce history and graph knowledge through concatenation with an input format as $U_{i-2}$ [SEP] $U_{i-1}$ [SEP] $U_i$ [SEP] \verb|OReact| tail [SEP] \verb|oEffect| tail.

Both of the above sampling steps use a threshold of $0.7$ between processed sub-utterances and matched heads, to reduce noise introducing of our sampled knowledge. The accuracies of baseline methods are reported in Table~\ref{proposed tasks}. $Base$ denotes only using the utterance to do prediction. $Knowledge$ and $History$ denote whether to add knowledge we sampled and dialogue history to the model. 
While adding knowledge improves the model performances, it seems problematic to directly concatenating history dialogues, which may bring noises. The moderate scores also indicate that there is still a room to improve for graph-grounded conversation understanding.

\section{Discussions of Future Work}
In this work, we provide a systematic approach from event mention detection, event linking to conversation graph construction which consists of 4 distinguished types of dialog flows. For each step, there exist possible refinements. For example, we plan to include other event-based resources to improve graph-conversation matching accuracy as well as the graph knowledge coverage. 

We also plan to continue the annotations to supply more dialog flow information especially those empathy ones, and evaluate more dialog flow relations on other datasets. 

\section*{Acknowledgements and Ethical Considerations}
We would like to thank the anonymous reviewers for their constructive comments. This work was approved by XiaoAI team. All personally identifiable information in our dataset was removed.

At last, we discuss the potential ethic impacts of this work. (1) \textbf{Transparency}: We will release the newly introduced corpus and the built conversation knowledge graph, as well as the benchmark approaches to facilitate future research. Similar datasets and knowledge bases include EmpatheticDialogues~\cite{Rashkin2019TowardsEO} and ATOMIC~\cite{atomic}, which are often public available and have been used extensively. (2) \textbf{Privacy}: The corpus is crowdsourced under a set of specific rules to forbid the workers disclosure sensitive and personal identifiable information. (3) \textbf{Politeness}: Because our conversations are human-written and are related to healthy dailylife scenarios, they are expected to be clean, legal, and polite. The crowdsourcing rules are designed to avoid emotionally triggering words as much as possible. 

\bibliography{custom}
\bibliographystyle{acl_natbib}

\newpage
\appendix

\section{Corpus: CConv}

\subsection{Example \& Statistics}
In our corpus CConv, conversations are conducted based on a scenario between two parties. Table~\ref{tab:corpus_example} gives an example conversation. The statistics of CConv is also present in Table~\ref{tab:corpus_stats}. Since there are 200 scenarios in total, and hence we have 160 diverse multi-turn conversations in average. 

\begin{table}[!h]
\centering
\begin{tabular}{lc} 
\hline
\# sessions of dialogues & 32,612 \\
\# utterances & 650,147 \\ 
\# unique scenarios & 200\\
\# conversation topics & 15 \\
Avg. \# words per utterance & 7.8\\
Avg. \# turns per dialogue & 19.9 \\
\hline
\end{tabular}
\caption{The Statistics of the Corpus CConv.}
\label{tab:corpus_stats}
\end{table}

\subsection{Topics and Scenarios}
To ensure the diversity of the conversations, we select 15 everyday topics. For each topic, we manually write tens of one-sentence scenario to guide the conversation context.

In total, we have 15 topics and 200 scenarios. To better understand, we show some example topics and scenarios in Table~\ref{tab:scenario_example}.

\begin{table*}[!h]
\centering
\begin{tabular}{|c|c|c|c|}
\hline
\multicolumn{4}{|c|}{Situation}\\\hline
\multicolumn{4}{|c|}{\begin{CJK*}{UTF8}{gkai}同事之间，一方身体不舒服，另一方表达关心\end{CJK*}} \\
\multicolumn{4}{|c|}{Acted as colleagues, one person is sick, and the other one cares about his/her health.}\\\hline
\multicolumn{4}{|c|}{Conversation}\\\hline
Speaker & Utterance & Emotion & Intent \\\hline
\multirow{2}{*}{1} & {\begin{CJK*}{UTF8}{gkai}你今天来得比平时晚呀。\underline{是身体不舒服吗}？\end{CJK*}} & \multirow{2}{*}{other} & \multirow{2}{*}{ask} \\
 & (You are later than normal days. Are you OK?)  &&\\
\hline
\multirow{3}{*}{2} & {\begin{CJK*}{UTF8}{gkai}呜呜，\underline{昨晚空调开的太大}，一大早起来\underline{头就特别疼}。\end{CJK*}} & \multirow{3}{*}{sad} & \multirow{3}{*}{description} \\
& (Yesterday the air conditioner was too cold that &&\\
& I had a headache this morning.) & & \\
\hline
\multirow{2}{*}{1} & {\begin{CJK*}{UTF8}{gkai}怪不得，那你\underline{吃过感冒药了吗}？\end{CJK*}} & \multirow{2}{*}{other} & \multirow{2}{*}{ask} \\
 & (I know. Have you taken the medicine?)  &&\\
\hline
\multirow{2}{*}{2} & {\begin{CJK*}{UTF8}{gkai}吃过了，现在已经好多了，就是有点想睡觉。\end{CJK*}} & \multirow{2}{*}{other} & \multirow{2}{*}{description} \\
 & (Sure. I feel better now, just feel a little bit sleepy.)  &&\\
\hline
...&...&...&...\\\hline
\multirow{2}{*}{2} & {\begin{CJK*}{UTF8}{gkai}今天的工作安排多么？\end{CJK*}} & \multirow{2}{*}{other} & \multirow{2}{*}{other} \\
 & (What are today's arrangements?)  &&\\
 \hline
\multirow{2}{*}{1} & {\begin{CJK*}{UTF8}{gkai}我会帮你做的。\underline{你好好休息吧}！\end{CJK*}} & \multirow{2}{*}{other} & \multirow{2}{*}{advise} \\
 & (I will help finish them. You'd better take a good rest.)  &&\\
 \hline
 \multirow{2}{*}{2} & {\begin{CJK*}{UTF8}{gkai}真是太感谢你了！\end{CJK*}} & \multirow{2}{*}{joy} & \multirow{2}{*}{other} \\
 & (I really appreciate a lot for your help!)  &&\\
 \hline
\end{tabular}
\caption{Example Conversation with Annotations. Note that the underlined words stand for the emotion cause span. Words are shorten due to space limit.}
\label{tab:corpus_example}       
\end{table*}

\begin{table*}[!h]
\centering
\begin{tabular}{|c|c|}
\hline
Topic & Scenario \\\hline
\multirow{4}{*}{Study} & {\begin{CJK*}{UTF8}{gkai}两个学生之间，讨论课业压力大，总是做不完作业\end{CJK*}} \\
& (Between two students, discuss the overload homework) \\
& {\begin{CJK*}{UTF8}{gkai}考研失败，向朋友倾诉自己的伤心和烦恼\end{CJK*}}\\
 & (Fail the entry exam of graduate study, express the distress to a friend)\\\hline
\multirow{4}{*}{Entertainment} & {\begin{CJK*}{UTF8}{gkai}讨论自己最喜欢的一部电影，以及为什么喜欢它\end{CJK*}} \\
& (Discuss one of your favorite films and why) \\
& {\begin{CJK*}{UTF8}{gkai}聊一聊自己曾经单曲循环过的歌曲，以及当时自己的感受\end{CJK*}}\\
 & (Talk about a music or a song you have put on repeat all the night)\\\hline
 \multirow{4}{*}{Love} & {\begin{CJK*}{UTF8}{gkai}情侣之间，因为生活作息不一致而吵架闹别扭\end{CJK*}} \\
& (Between a couple, quarrel with the lover due to inharmonious habits) \\
& {\begin{CJK*}{UTF8}{gkai}自己订婚了，激动地与好友分享喜讯\end{CJK*}}\\
 & (Being engaged, share the good news to the best friend)\\\hline
\end{tabular}
\caption{Example Topics and Scenarios.}
\label{tab:scenario_example}       
\end{table*}

\subsection{Annotation Criteria}

To facilitate future research, we hire another 3 well-trained assistants to manually annotate the conversations with fine-grained emotional labels including speaker's emotion type, emotion cause, and response intention type. The annotation example in given along with the example in Table~\ref{tab:corpus_example}.

\noindent\textbf{Emotion Class}. Following~\citet{Rashkin2019TowardsEO}, we define emotion type with 5 general classes \{joy, angry, sad, surprising, other\}. 

\noindent\textbf{Emotion Cause Span}. Emotion cause span is a continuous text spans implying the reason of certain emotion~\cite{li2021towards}.

\noindent\textbf{Response Intent}. Response intention type is essential for building empathetic chatbots, and we define 6 commonly-adopted intent classes of \{ask, advise, describe, opinion, console, other\} following~\citet{Welivita2020ATO}, which are described in Table~\ref{tab:intent}.

\begin{table*}[!h]
\centering
\begin{tabular}{|l|l|l|} 
\hline
Intent Type & Definition & Example \\
\hline
ask & to know further details or
clarify & \emph{What happended?} \\
\hline
describe & present more details and explain the reasons & \emph{I'm sad because I failed the exam.} \\\hline
advise & give explicit solutions & \emph{Try to exercise more.} \\\hline  
opinion & share own thoughts & \emph{I don't like being disturbed after work}.\\\hline
console & pacify others & \emph{I hope you'd feel better.}\\\hline
other & - & \emph{Goodbye.} \\\hline
\end{tabular}
\caption{Annotation Criteria for Response Intent.}
\label{tab:intent}
\end{table*}

\section{ATOMIC}
In this work, we introduce ATOMIC~\cite{atomic} as the commonsense knowledge base due to its attractive properties of mental state inferences and \emph{if-then} causal relations, as analyzed before. 

ATOMIC~\cite{atomic} is a novel event-centered knowledge graph, consisting of 880K tuples of social commonsense knowledge. Distinguished from ConceptNet~\cite{Speer2017ConceptNet5A}, there are two unique properties making ATOMIC suitable and attractive for building empathic chatbots. Firstly, ATOMIC collects knowledge about how people will feel and react to a given event. This kind of knowledge is related to people's mental states, which is beneficial for understanding implicit emotions. For example, given a head event \emph{PersonX makes PersonY’s coffee}, ATOMIC contains knowledge that PersonY will be \emph{grateful} along the relation \verb|oReact|. 
Secondly, ATOMIC organizes knowledge using several inferential relations and naturally supports \emph{if-then} reasoning, which is crucial generating coherent responses. 

\begin{figure*}[th!]
    \centering
    \includegraphics[width=0.99\textwidth]{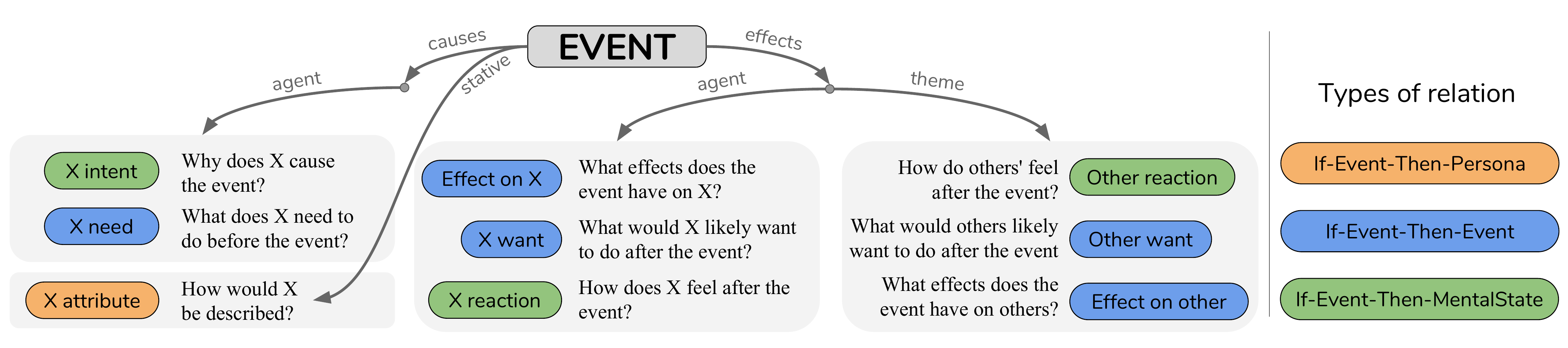}
    \caption{
    The taxonomy of \textit{if-then} reasoning types. We consider nine \emph{if-then} relations that have overlapping hierarchical structures as visualized above.
    One way to categorize the types is based on the type of content being predicted: (1) \textbf{If-Event-Then-Mental-State}, (2) 
\textbf{If-Event-Then-Event}, and (3) \textbf{If-Event-Then-Persona}.
    Another way is to categorize the types based on their causal relations: (1) \textbf{``causes''}, (2) \textbf{``effects''}, and (3) \textbf{``stative''}.
    Some of these categories can further divide depending on whether the reasoning focuses on the ``agent'' (X) or the ``theme'' (Other) of the event.
    }
    \label{fig:inference-dimensions}
\end{figure*}

Here, we adopt the figures and demonstrations from the original ATOMIC paper~\cite{atomic} to present the 9 relations defined in ATOMIC and give some examples in Figure~\ref{fig:inference-dimensions} and Table~\ref{tab:annotation-examples}.

\begin{table*}[!htb]
\footnotesize
\centering
\resizebox{\textwidth}{48mm}{
\begin{tabular}{@{}llll@{}}
\toprule
Event                                              & Type of relations       & Inference examples                                                                                                                           & Inference dim. \\ \midrule
\multirow{7}{*}{\begin{tabular}[c]{@{}l@{}}``PersonX pays PersonY\\a compliment''\end{tabular} } & If-Event-Then-Mental-State & \begin{tabular}[c]{@{}l@{}}PersonX wanted to be nice\\ PersonX will feel good\\ PersonY will feel flattered\end{tabular}                   & \begin{tabular}[c]{@{}l@{}}xIntent\\ xReact\\ oReact\end{tabular} \\ \cmidrule(l){2-4} 
                                                   & If-Event-Then-Event       & \begin{tabular}[c]{@{}l@{}}PersonX will want to chat with PersonY\\ PersonY will smile\\ PersonY will compliment PersonX back\end{tabular} & \begin{tabular}[c]{@{}l@{}}xWant\\ oEffect\\ oWant\end{tabular} \\ \cmidrule(l){2-4} 
                                                   & If-Event-Then-Persona     & \begin{tabular}[c]{@{}l@{}}PersonX is flattering\\ PersonX is caring\end{tabular}                                                          & \begin{tabular}[c]{@{}l@{}}xAttr\\ xAttr\end{tabular}\\ \midrule
\multirow{7}{*}{\begin{tabular}[c]{@{}l@{}}``PersonX makes \\ PersonY's coffee''\end{tabular}}    & If-Event-Then-Mental-State & \begin{tabular}[c]{@{}l@{}}PersonX wanted to be helpful\\ PersonY will be appreciative\\ PersonY will be grateful\end{tabular}             & \begin{tabular}[c]{@{}l@{}}xIntent\\ oReact\\ oReact\end{tabular} \\ \cmidrule(l){2-4} 
                                                   & If-Event-Then-Event       & \begin{tabular}[c]{@{}l@{}}PersonX needs to put the coffee in the filter\\ PersonX gets thanked\\ PersonX adds cream and sugar \end{tabular}  & \begin{tabular}[c]{@{}l@{}}xNeed\\ xEffect\\ xWant\end{tabular} \\ \cmidrule(l){2-4} 
                                                   & If-Event-Then-Persona     & \begin{tabular}[c]{@{}l@{}}PersonX is helpful\\ PersonX is deferential\end{tabular}                                                        & \begin{tabular}[c]{@{}l@{}}xAttr\\ xAttr\end{tabular}\\ \midrule
\multirow{7}{*}{``PersonX calls the police''}      & If-Event-Then-Mental-State & \begin{tabular}[c]{@{}l@{}} PersonX wants to report a crime \\ Others feel worried \end{tabular}                 & \begin{tabular}[c]{@{}l@{}}xIntent\\ oReact\end{tabular} \\ \cmidrule(l){2-4} 
                                                   & If-Event-Then-Event       & \begin{tabular}[c]{@{}l@{}}PersonX needs to dial 911 \\ PersonX wants to explain everything to the police\\ PersonX starts to panic \\ Others want to dispatch some officers \end{tabular} & \begin{tabular}[c]{@{}l@{}}xNeed\\ xWant\\ xEffect \\ oWant \end{tabular} \\ \cmidrule(l){2-4} 
                                                   & If-Event-Then-Persona     & \begin{tabular}[c]{@{}l@{}}PersonX is lawful\\ PersonX is responsible\end{tabular}                                                        & \begin{tabular}[c]{@{}l@{}}xAttr\\ xAttr\end{tabular}\\ \bottomrule
\end{tabular}}
\caption{Examples of \textbf{If-Event-Then-X} commonsense knowledge present in~\citet{atomic}.
For inference dimensions, ``x'' and ``o'' pertain to PersonX and others, respectively (e.g., ``xAttr'': attribute of PersonX, ``oEffect'': effect on others).}
\label{tab:annotation-examples}
\end{table*}

\section{Translation Method}

\subsection{Replacement of Certain Tokens}
We begin with translating high-frequency patterns in the original triplets. As compared to the pre-defined set of relations, it is more difficult to handle the heads and tails. In ATOMIC, for example, there exist \textbf{185,046} heads and tails containing tokens like ``\emph{PersonX}'' and ``\emph{PersonY}''. These personal pronouns stand for the givers and the receives for a certain event, and can be regarded as the speech parties in a conversation. Also, some ATOMIC heads like \{\emph{PersonX gets \_\_\_\_ as a pet}\}, have a blank which can be filled with various tokens. 

These aforementioned patterns bring ambiguity to the triplet semantics, and will confuse the translation model. To address, we devise a series of replacement rules to keep the original semantics while translation. For example, for the ATOMIC head \emph{PersonX votes for personY}, we convert it to be ``Someone votes for someone else'' and send it to our translation model.

\subsection{Joint Translation of Head and Tail}
Nevertheless, the majority of the heads and tails in ATOMIC are short phrases, while machine translation models are often context-based. The multi-sense characteristics of language will further deteriorate the translation quality if we separately feed each single head and tail to a translation model.

To remedy the issues, we instead translate the head and tail in each triplet together. Given a triplet <$h,r,t$>, we connect the head $h$ with its $t$ using a heuristic connecting word $r'$ w.r.t. the relation $r$, and obtain one long sentence $l$. After translating the long text, we split the translation result with the connecting word and turn it into $h_{tr}$ and $t_{tr}$:
\begin{equation}
\begin{aligned}
l &= \mathrm{CONNECT}(h, r', t)\\
l^{'}_{tr} &= \mathrm{TRANSLATION}(l)\\
h_{tr}, r_{tr}, t_{tr} &= \mathrm{SPLIT}(m^{'}_{tr}, r'_{tr})
\end{aligned}
\end{equation}
where the resulted <$h_{tr}$, $r_{tr}$, $t_{tr}$> is the translated triplets. And $\mathrm{CONNECT}$, $\mathrm{SPLIT}$ denote the corresponding operation. $\mathrm{TRANSLATION}$ stands for our translation model. By this means, we expect the connected $l$ provides more contextual information for better semantic translation. The comparison results between separate translation and joint translation will be given in Section~\ref{sec:node_evaluation}.  

Note that auxiliary translation methods can be used. In this work, we use Xiaomi commercial Translation service.\footnote{\url{http://fanyi.mioffice.cn}} For simplicity, we denote the translated ATOMIC as \textbf{ATOMIC-zh}.

\section{Evaluation}
\subsection{Template-based Event Extraction Methods}

To evaluate our matching methods proposed in this work, we randomly choose 100 utterances and compare with several approaches. In specific, we propose a baseline $POS$ matching method, which employs POS tagging-based templates to extract events. The templates are given in Table~\ref{tab:template}.

\subsection{More Examples of Constructed Scenario Graphs and Annotation Tool}
\begin{figure}[htbp]
    \centering
    \includegraphics[width=7.7cm]{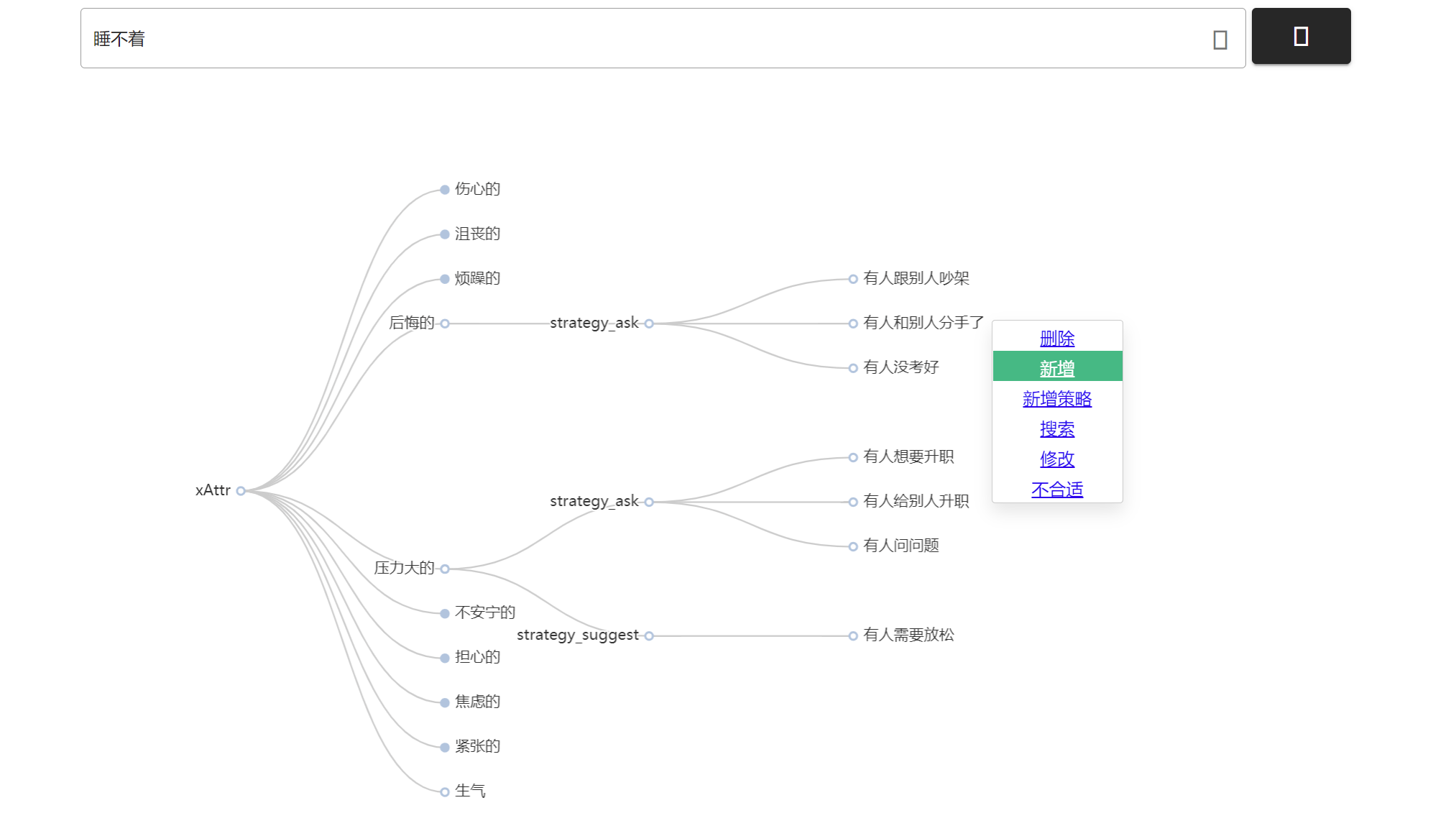}
    \caption{Adding Tails Function in Our Annotation Tool.}
    \label{fig:tool1}
\end{figure}

\begin{figure}[htbp]
    \centering
    \includegraphics[width=7.7cm]{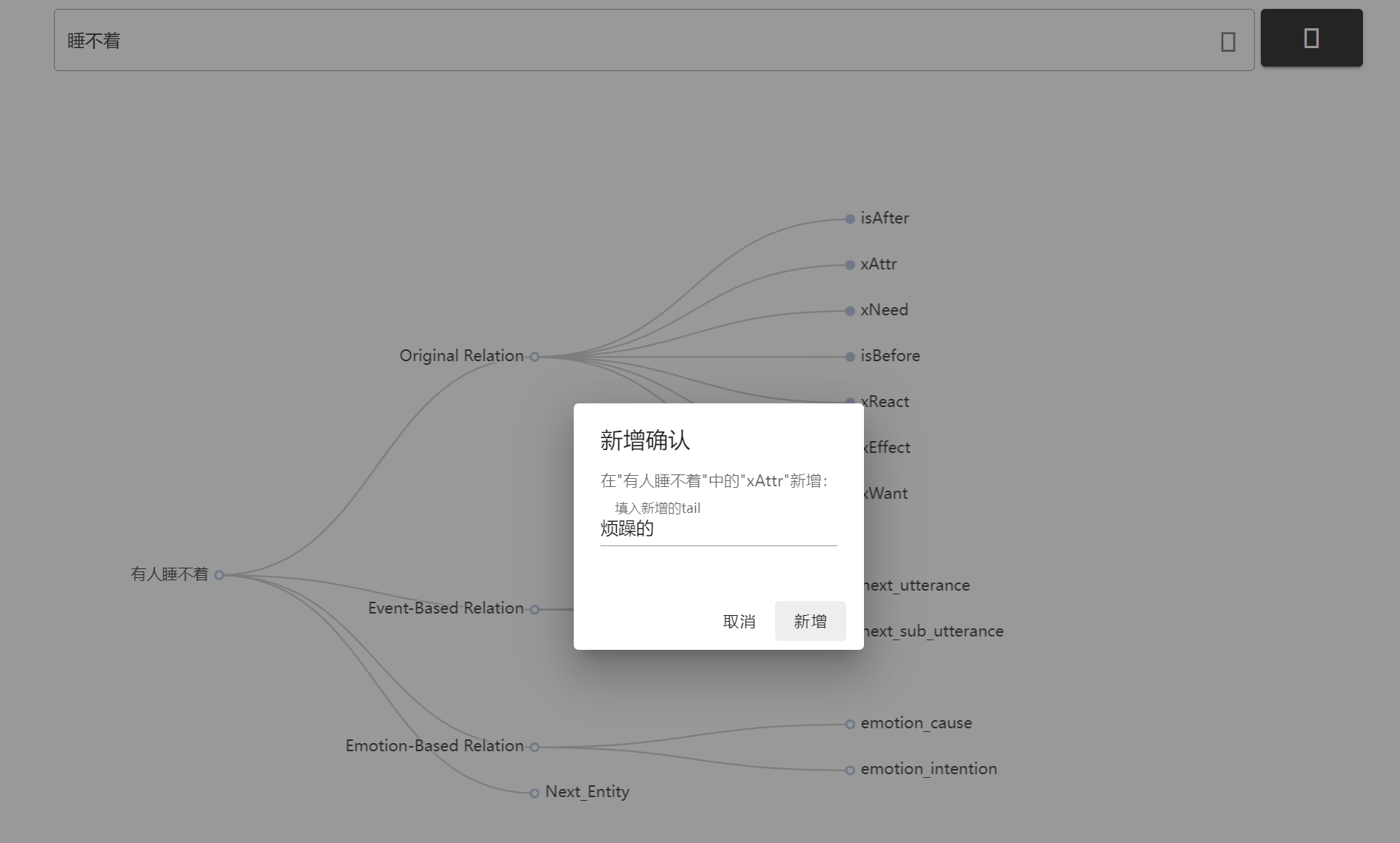}
    \caption{Adding Tails Function in Our Annotation Tool.}
    \label{fig:tool2}
\end{figure}

In this section, we visualize more snippets of the scenario graphs. They are ``insomnia'' in Figure~\ref{fig:insomnia}. We also give examples of revising function in our interactive annotation tool in Figure~\ref{fig:tool1} and Figure~\ref{fig:tool2}, with the head \begin{CJK*}{UTF8}{gkai}``有人睡不着''\end{CJK*} (someone cannot fall asleep).

Please kindly note that for clarity, we only visualize a small set of relation and tails in each figure, and try to give a comprehensive view of the relations by showing different relations in different scenario graphs. In fact, every scenario graphs contain the full set of C$^3$KG relations. 

\begin{figure}[htbp]
    \centering
    \includegraphics[width=7.7cm]{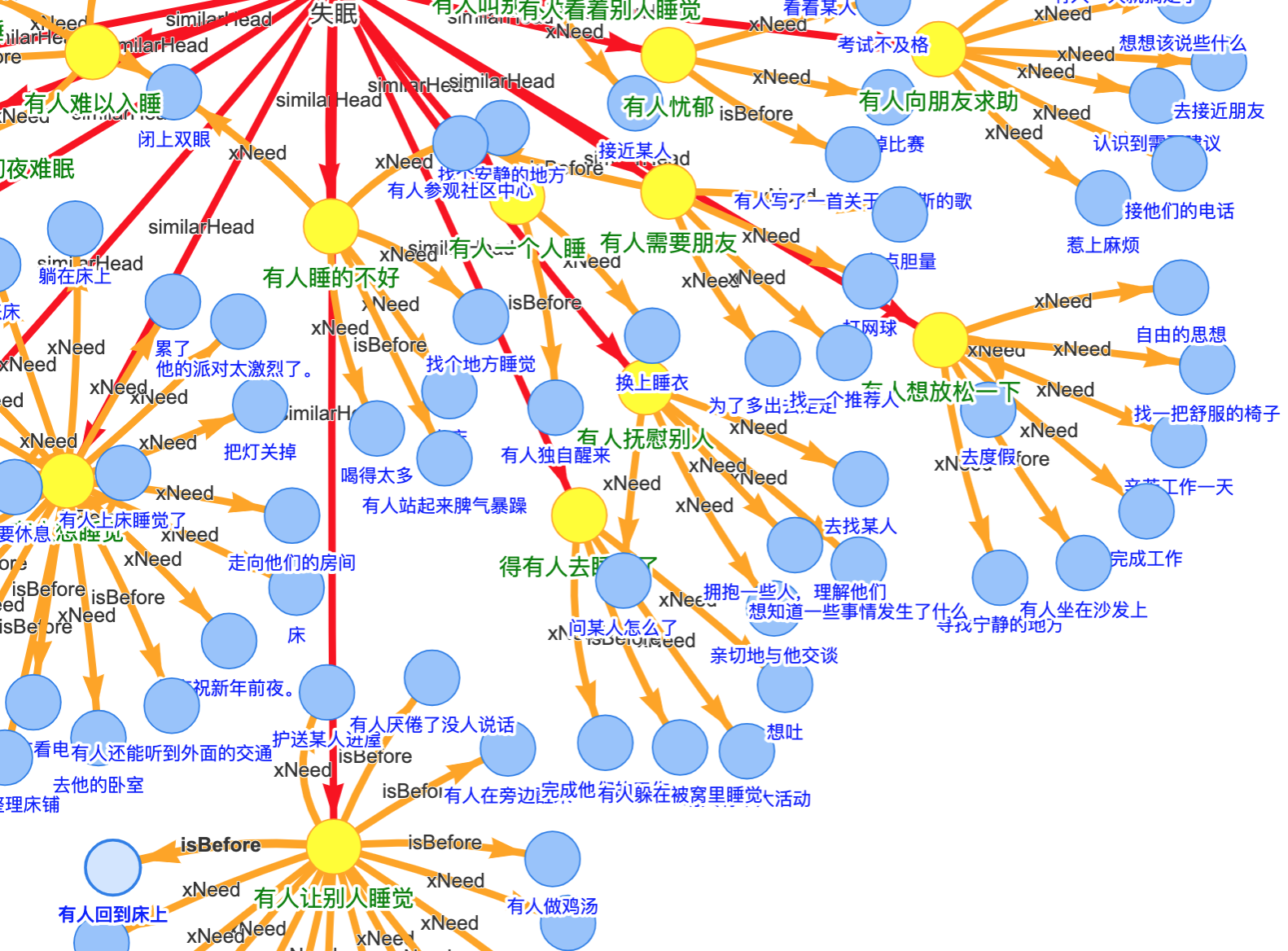}
    \caption{Scenario Graph of ``Insomnia''.}
    \label{fig:insomnia}
\end{figure}

\begin{figure}[htbp]
    \centering
    \includegraphics[width=7.7cm]{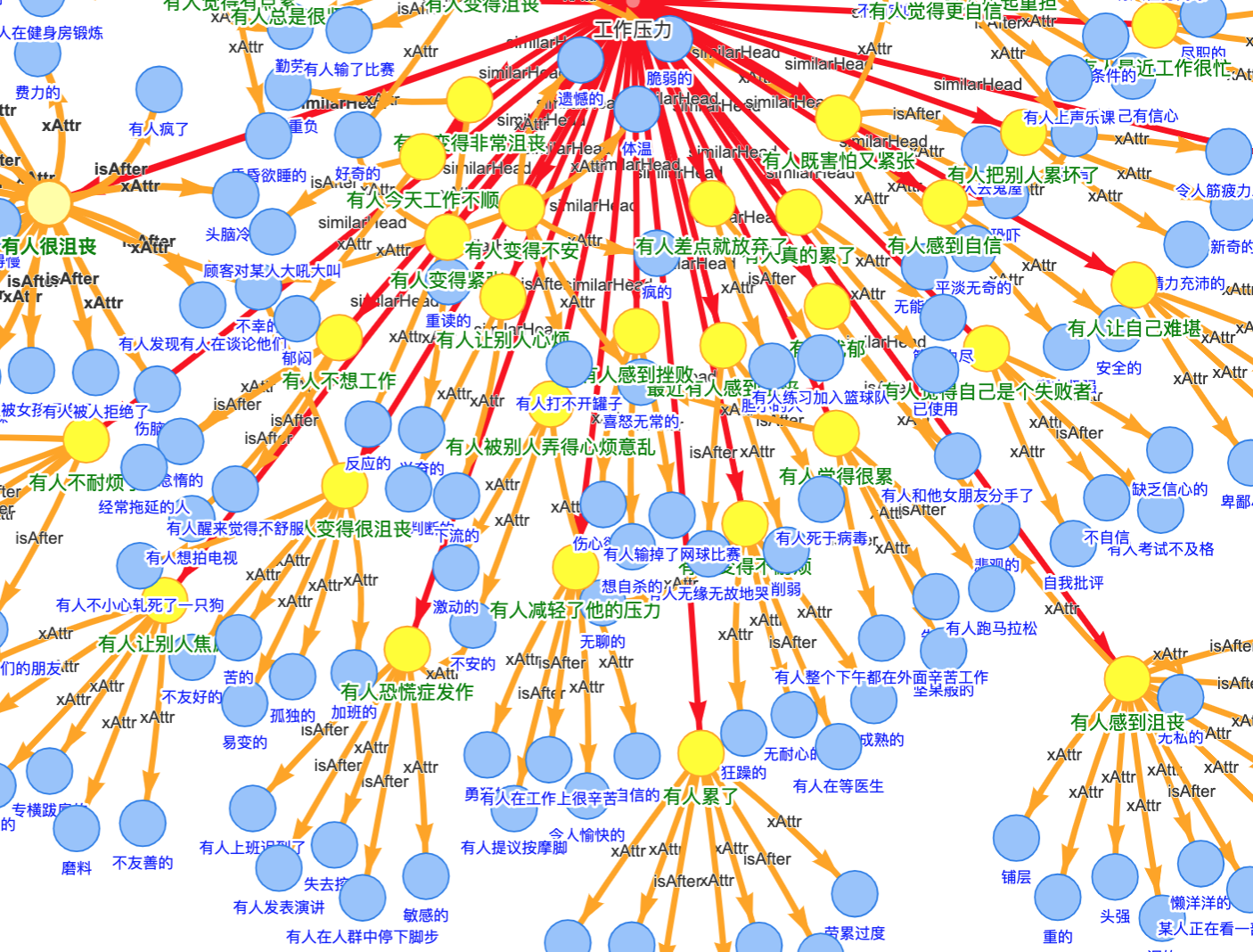}
    \caption{Scenario Graph of ``Work Pressure''.}
    \label{fig:insomnia}
\end{figure}

\begin{table}[]
\begin{tabular}{@{}cc@{}}
\toprule
\multicolumn{1}{l}{POS sequence} & Example              \\ \midrule
v+v &   {\begin{CJK*}{UTF8}{gkai}想睡觉\end{CJK*}} (want to sleep) \\
v+n &  {\begin{CJK*}{UTF8}{gkai}做作业\end{CJK*}} (do homework)\\
v+i &  {\begin{CJK*}{UTF8}{gkai}感觉如释重负\end{CJK*}} (feel relieved)                    \\
v+u+z &  {\begin{CJK*}{UTF8}{gkai}跑得飞快\end{CJK*}} (run fast) \\
v+u+m &  {\begin{CJK*}{UTF8}{gkai}看了一下\end{CJK*}} (take a look)\\
v+c+v &  {\begin{CJK*}{UTF8}{gkai}讨论并通过\end{CJK*}} (discuss and approve)                    \\
v+c+i &  {\begin{CJK*}{UTF8}{gkai}尝试但一无所获\end{CJK*}} (try but find nothing)                    \\
a+v  &  {\begin{CJK*}{UTF8}{gkai}热烈鼓掌\end{CJK*}} (applause warmly)                    \\              
\bottomrule
\end{tabular}
\caption{POS templates we use in event extraction method $POS$.}
\label{tab:template}
\end{table}

\begin{table}[]
\begin{tabular}{@{}cc@{}}
\toprule
Original pattern                             & Replaced pattern                                \\ \midrule
PersonX...PersonX...   & Someone...himself...      \\ 
PersonX...PersonY...   &Someone...some one else... \\ 
PersonX...PersonX's... & Someone...his...           \\ 
PersonX...PersonY's... & Someone...someone else's   \\ 
...\_\_\_... & ...something... \\
\bottomrule
\end{tabular}
\caption{Pattern replacement we use when translating ATOMIC}
\end{table}

\end{document}